\documentclass[conference]{IEEEtran}
\IEEEoverridecommandlockouts
\usepackage{cite}
\usepackage{float}
\usepackage{tfrupee}
\usepackage{amsmath,amssymb,amsfonts}
\usepackage{algorithmic}
\usepackage{cite}
\usepackage{csquotes}
\usepackage{multirow}
\usepackage{graphicx}
\usepackage{textcomp}
\usepackage{tikz}
\usetikzlibrary{shapes, arrows,positioning}
\usepackage{xcolor}
\usepackage{caption}
\usepackage{subcaption}
 \usepackage{lipsum}
 \usepackage{graphicx}
\def\BibTeX{{\rm B\kern-.05em{\sc i\kern-.025em b}\kern-.08em
    T\kern-.1667em\lower.7ex\hbox{E}\kern-.125emX}}
\begin{document}

\title{An Autonomous Hybrid Drone-Rover Vehicle for Weed Removal and Spraying Applications in Agriculture\\
}
	\author{
	\IEEEauthorblockN{J Krishna Kant}
		\IEEEauthorblockA{\textit{Student} \\
			\textit{IIITDM Kancheepuram}\\
			Chennai, India \\
			evd19i001@iiitdm.ac.in
			}
		\and	
		\IEEEauthorblockN{Mahankali Sripaad}
		\IEEEauthorblockA{\textit{Student} \\
			\textit{IIITDM Kancheepuram}\\
			Chennai, India \\
			mfd19i001@iiitdm.ac.in
		}
			\and
   		\IEEEauthorblockN{Anand Bharadwaj}
		\IEEEauthorblockA{\textit{Student} \\
			\textit{IIITDM Kancheepuram}\\
			Chennai, India \\
			mpd19i002@iiitdm.ac.in
			}
		\and
		\IEEEauthorblockN{Rajashekhar V S}
		\IEEEauthorblockA{\textit{ Project Engineer} \\
			\textit{AIRL,IISc}\\
			Bangalore, India \\
			vsrajashekhar@gmail.com
			}

		\and
		\IEEEauthorblockN{Dr. Suresh Sundaram}
		\IEEEauthorblockA{\textit{Dept of Aerospace} \\
			\textit{Assoc. Prof IISc}\\
			Bangalore, India \\
			vssuresh@iisc.ac.in
			}
		\and

}
\maketitle
\begin{abstract}
The usage of drones and rovers helps to overcome the limitations of traditional agriculture which has been predominantly human-intensive, for carrying out tasks such as removal of weeds and spraying of fertilizers and pesticides. Drones and rovers are helping to realize precision agriculture and farmers with improved monitoring and surveying at affordable costs. Major benefits have come for vertical farming and fields with irrigation canals. However, drones have a limitation of flight time due to payload constraints. Rovers have limitations in vertical farming and obstacles like canals in agricultural fields. To meet the different requirements of multiple terrains and vertical farming in agriculture, we propose an autonomous hybrid drone-rover vehicle that combines the advantages of both rovers and drones. The prototype is described along with experimental results regarding its ability to avoid obstacles, pluck weeds and spray pesticides. 
\end{abstract}


\section{Introduction}
The growing evolution in electronics, image processing sensors, automation, rovers, and drones has enabled many agricultural tasks such as spraying fertilizers and pesticides, to be mechanized and automated to overcome the shortage of rural labour, caused due to migration of rural labour to urban areas with an increase in urbanization. 

The works of \cite{ribeiro2022computational, rajeshwari2021smart} describe autonomous rovers which remove weeds by spraying herbicides and by cutting them respectively, with the help of image processing and Artificial Intelligence. The work of \cite{quaglia2020agri} describes a solar-powered robot that is used to monitor crops, leaves, and soil. The work of \cite{velhal2022decentralized} describes the concept of task allocation to defenders guarding a territory which can be co-related to the effective utilization of available rovers and drones of a particular farm. The works of \cite{kulkarni2023heli, sharma2023geometric, kumar2009direct} describe drone literature, related to payload and stability analysis. The work of \cite{esposito2021drone} makes use of machine-learning techniques to identify weed patches.
 \begin{figure}[!h]
    \centering
    \includegraphics[scale=0.2]{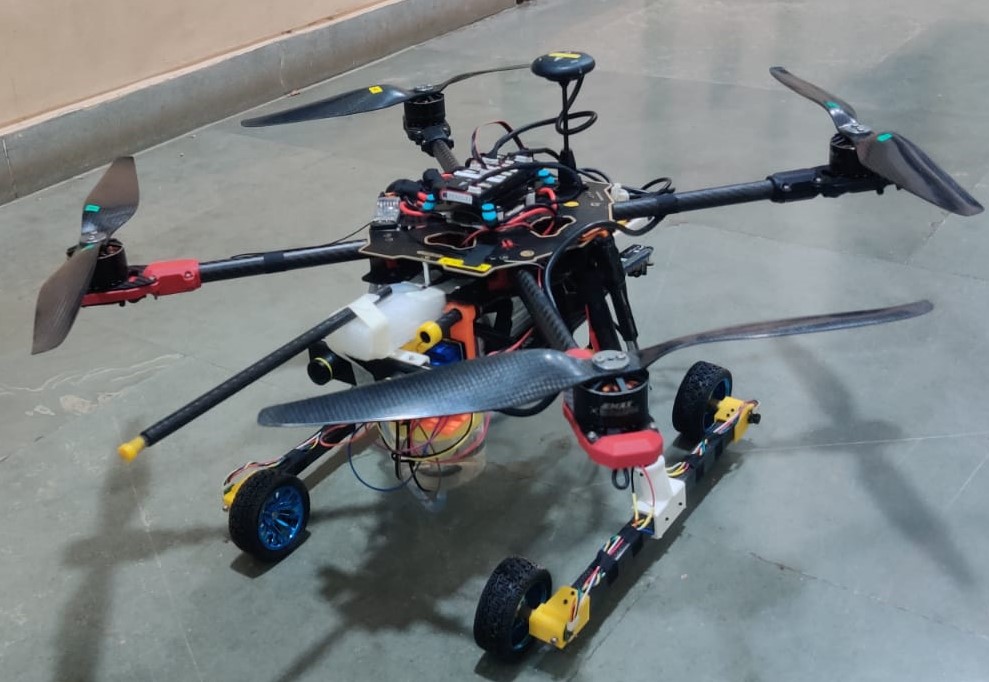}
    \caption{Prototype of Hybrid Vehicle with sprayer arrangement}
    \label{fig_intro}
\end{figure}

Rovers have better efficiency compared to drones for ground-based operations. Drones can cover a larger area in a lesser amount of time. In a few farms such as potatoes and grapes, there is a need to tackle the weeds at the stem level which can be done with rovers. In farms, there are canals and uneven terrains. Drones can easily fly over them, unlike rovers. Drones can also easily operate in vertical farming where the crops are present at different levels. Hence, there is a need to have a hybrid vehicle that helps to combine the benefits of both drones and rovers. The works of \cite{esposito2021drone, mammarella2022cooperation} describe the importance of a hybrid vehicle and the benefits associated with a symbiotic relationship between rover and drone. Hybrid vehicle designs are described in \cite{asadi2020integrated, daler2015bioinspired, abddrift}. A hybrid vehicle that can fly and walk is seen in \cite{daler2015bioinspired}. In the works of \cite{asadi2020integrated} and \cite{abddrift}, hybrid vehicle designs where the rover and drone can be decoupled are described, for indoor construction site data management and fire tracking respectively. 

The paper describes the development of a compact hybrid drone rover vehicle with sprayer and plucker arrangement, respectively. The proof of concept has been built using a 650mm drone Tarot frame, which is smaller than the regular agricultural drones of frame size greater than or equal to 960mm. This is shown in Figure \ref{fig_intro}. The paper is organized as follows. Section II describes the design methodology flow, from system requirements to proof-of-concept development. Section III describes the experimental results obtained. They are used to validate the proposed features of the developed prototype of the hybrid vehicle, such as obstacle detection, the transition between ground and flight mode in the presence of an obstacle, weed detection, and tackling them using plucker and sprayer arrangement. Section IV gives the conclusion.
\section{Methodology}
The following section describes the development of the prototype along the lines of the design methodology. It consists of the process of understanding the needs of the customer (Empathize), identifying the system requirements (Define), developing the conceptual design (Ideation), building the prototype, and validating it (Test).
\subsection{Problem Definition}
\subsubsection{System Requirements}
There is a need to have a compact hybrid vehicle for increased mobility and flexibility. There is also a need to have a lightweight weed removal system which is able to tackle different types of weeds in farms. It should not affect the flight time of the system due to its weight. There is a need to have a system which does the tasks autonomously to overcome the reduced availability of rural labour which has become expensive in recent times.
\subsection{Conceptual Design} 
\begin{figure}[htp]
\begin{center}
\includegraphics[scale=0.250]{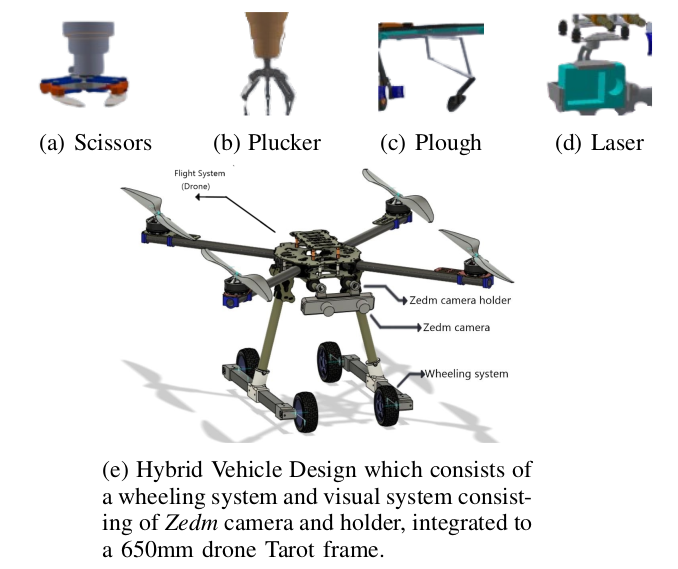}
\end{center}
\caption{Conceptual Design of hybrid drone-rover vehicle and weed removal methods}
\label{fig_concept}
\end{figure}

This section describes the modeled system and its features. The modeled design of the hybrid drone-rover vehicle and the various weed removal methods are indicated in Figure \ref{fig_concept}.

\subsubsection{Selection of Landing Gear}
A cuboid-shaped landing gear is preferred over the conventional cylindrical landing gear, as the former is stronger than the latter and the motors can be clamped easily.
\begin{equation}
Stiffness \propto EI/L{^3}
\end{equation}
\begin{equation}
Relative\hspace{0.1cm} Shape\hspace{0.1cm} Factor = \sigma_{S} / \sigma_{C}  = Is/Ic 
\end{equation}
\begin{displaymath}
= (1/6 H{^3} (H-h) (1 + 3B/H)) / (\pi/4 ((d/2){^4} - (d1/2){^4}))
\end{displaymath} 

To find the relative strength between the two rods, we calculate the relative shape factor for the two rods, of same length ``L" (30 cm) and material.
The relative shape factor for the hollow square rod to the hollow circular rod is defined as Is/Ic as seen by the above set of equations. The bending stiffness of the hollow square rod is seen to be relatively 70\% better than that of a hollow circular rod. 

\begin{figure}[htp]
\begin{center}
\includegraphics[scale=0.250]{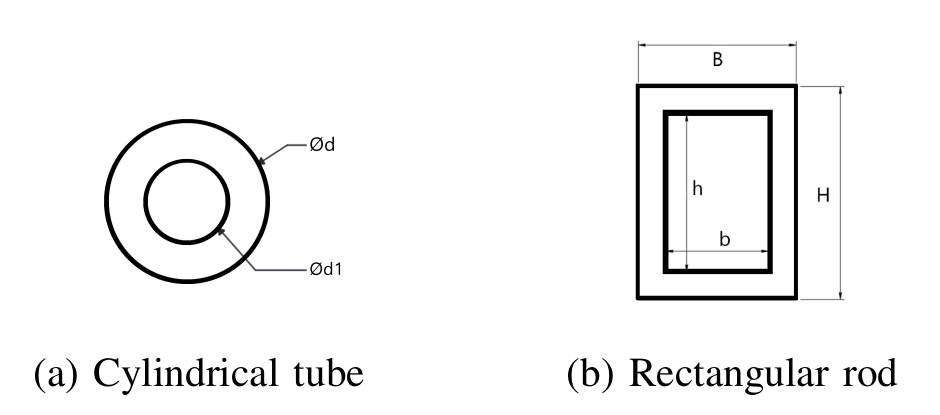}
\end{center}
\caption{Cross Section View of Landing Gear designs used to find the relative strength between the two designs.}
\label{fig_cross_section}
\end{figure}

The stiffness of the square (cuboidal) and circular (cylindrical) shaped rod is indicated by ``$\sigma_{S}$" and ``$\sigma_{C}$" respectively. Their respective Moment of Inertia is indicated by ``$I_{s}$" and ``$I_{c}$" respectively. ``E" indicates the Young's modulus of the rod.
The outer and inner diameters of the rod are 14 mm and 12 mm, as represented by \enquote{d} and \enquote{d1} respectively. \enquote{h} and \enquote{b} represent the length and breadth of the inner rectangle, whereas \enquote{H} and \enquote{B} represent the length and breadth of the outer rectangle, as shown in the reference image in Figure \ref{fig_cross_section}. For our system, we have considered a square rod. As a result, \enquote{h} and \enquote{b} are equal. \enquote{H} and \enquote{B} are also equal. The outer and inner sides of the rod are 14 mm and 12 mm respectively.
\subsubsection{Motor holder designs}
\begin{figure}[htp]
    \centering
    \includegraphics[scale=0.22]{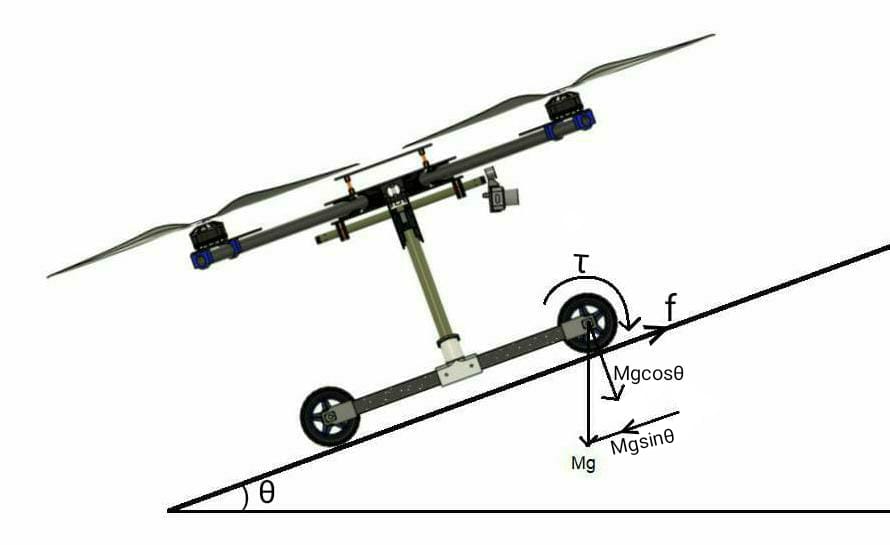}
    \caption{Free Body Diagram of the hybrid drone-rover vehicle over an inclined surface.}
    \label{fig_free_body}
\end{figure}

n-20 motors and BO motors were used for the two prototypes. In prototype-1, BO motors of torque 3.5 kg-cm and 150 RPM ratings were utilized. For prototype-2, n-20 motors which have a mass of 11g attached to an external gearbox with a ratio of 250:1, having a stall torque of 16 kg-Cm and a no-load speed of 120 RPM
were used. With the usage of n-20 motors, the motor holder design became more compact and weight reduced by a factor of fifty percent (about 10g) in comparison to design 1 of BO-motor holders. The BO-motor holder design weighs about 27g. The higher torque capability of n-20 motors helps to move on uneven terrains. The below set of equations has been used to calculate the specification of motors used in Prototype-2, where the vehicle can move on surfaces with an inclination of 15 \textdegree. Figure \ref{fig_free_body} describes the free-body diagram of the hybrid drone rover vehicle used for describing the below set of equations.
\begin{equation}
\Sigma Forces = f_{total}= \tau /r + f - f_{g} = Ma
\end{equation}
\begin{equation}
\tau  = M(a + g\sin \theta)r - f
\end{equation}
``r" is the radius of wheels of hybrid vehicle moving over a surface inclined at an angle ``$\theta$". ``a" is the acceleration of the hybrid vehicle. The acceleration due to gravity is indicated by ``g". The mass of the vehicle is indicated by ``4M". $``\tau"$ is the torque of the motor required. ``f" is the frictional force on the inclined surface.
\subsubsection{Visual System Integration}
Our vehicle makes use of a \emph{Zedm} camera with a resolution of 3840 x 1080p to carry out obstacle detection and the detection of weeds. The camera is attached to the vehicle with the help of the attachments as illustrated in part e of Figure \ref{fig_concept}.
\subsubsection{Weed Removal Methods}
The scissor-cut mechanism and plucker arrangement have telescopic arms to reach the bottom of the plant. The plough arrangement requires more power and might affect the top layer of soil. It might affect the stability of the drone during flight. Laser placed along with the gimbal makes use of point and shoot mechanism. Movement of the gimbal helps to shoot at the required location. Safety is a matter of concern for both laser and scissor-cut mechanisms. Plucker design is simple and easy to implement but spraying was considered to be a viable option for weed removal when extended to applications like vertical farming. The various weed removal methods considered are described in Figure \ref{fig_concept}.
\subsubsection{Overview of Design}
In the modeled system, the integration of a visual system along with the usage of square landing gear rods for the wheeling system was implemented. In developing the design, we considered a 650 mm Tarot drone frame. The frame size is smaller than that of regular agricultural drones with a frame size greater than 960 mm. Refer to part e of Figure \ref{fig_concept} for the hybrid vehicle design.
\subsection{System Design and Algorithms}
\subsubsection{System Overview}
\begin{figure}[htp]
    \centering
    \includegraphics[scale=0.23]{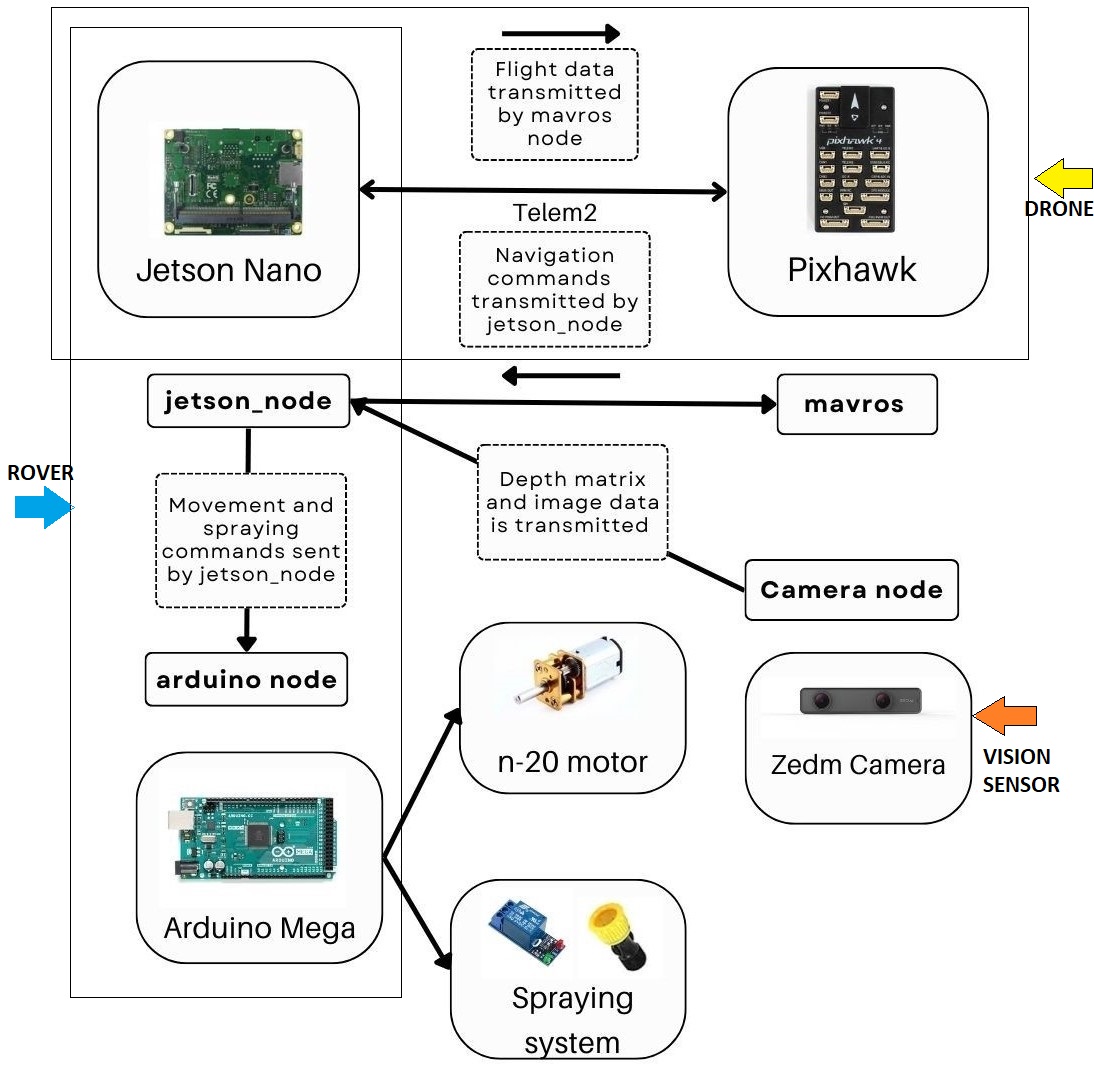}
        \caption{System Framework which describes all the nodes and their connectivity. The ``jetson\textunderscore node" is the master node that controls the operation between rover (``arduino node")  and drone mode (``mavros node") and spraying activity (``arduino node")}
    \label{fig_frame_work}
\end{figure}

The system works on the \emph{ROS} framework consisting of the following nodes. Figure \ref{fig_frame_work} represents the system framework. 
\begin{itemize}
    \item ``jetson\_node" (\emph{Jetson Nano})
    \begin{itemize}
     \item It controls the operation of the system between rover and flight modes.
     \item It also gives commands to spray on weeds.
  \end{itemize}
    \item ``Camera node" (\emph{Zedm})
    \begin{itemize}
     \item It sends Point cloud data (depth matrix) and image data feed to \emph{Jetson Nano}.
  \end{itemize}
     \item ``mavros node" (\emph{Pixhawk})
   \begin{itemize}
     \item It transmits IMU data which is taken by \emph{Jetson Nano} to enable offboard flight mode.
   \end{itemize}
      \item ``arduino node" (\emph{\emph{\emph{\emph{\emph{Arduino}}}} Mega})
   \begin{itemize}
     \item It is used to run the motors (rover mode) and sprayer system which is equipped with a relay module for timing operation.   
   \end{itemize}
\end{itemize}

\begin{figure*}[h!]
      \centering
    \includegraphics[scale=0.1]{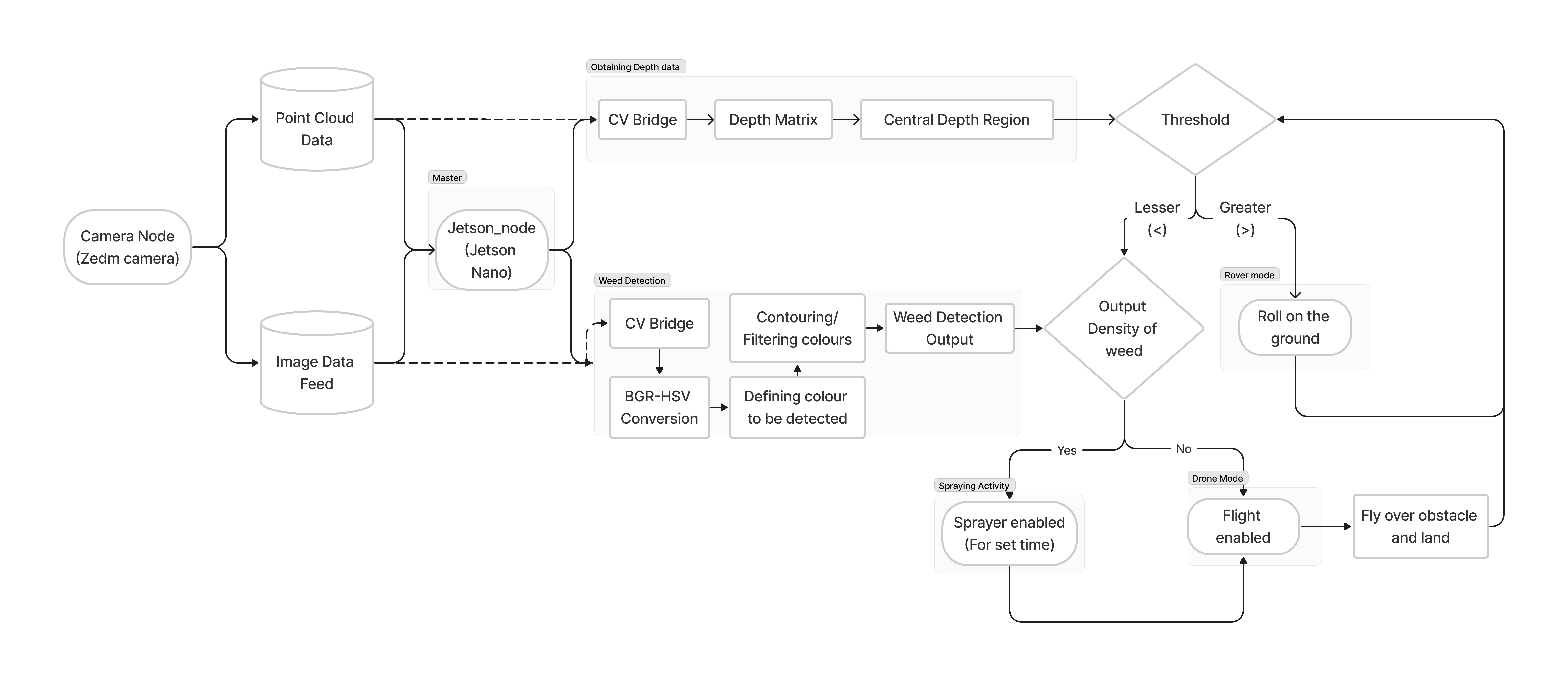}
    \caption{Overall working process of the system:-
    \\The hybrid vehicle continues to roll on the ground in the absence of obstacles. In the presence of an obstacle, it stops and tries to detect the presence of weed. The presence of an obstacle is detected based on the depth values received from the Camera node. The detection of weeds is based on HSV detection. If weed is not detected, it shifts to drone mode and flies over the obstacle. If weed is detected, it sprays and flies over it. It then continues to roll on the ground. The weed has been considered to be a crop placed in a pot. 
}
    \label{fig:Overall}
\end{figure*}
\subsubsection{System Working Principle}
The system has been designed for the situation where the vehicle rolls in the space present between the rows of crops and the weed is present in the path between rows of crops. For experimental purposes, the weed has been considered to be a crop placed in a pot. The details of the experiment are specified in the experimental results section. There are two aspects in the overall working of the hybrid vehicle. They are ground and aerial movement and detection and tackling of weeds.

The hybrid vehicle is equipped with a front camera (\emph{Zedm}). The image data feed and depth matrix data are sent by the Camera node (\emph{Zedm}) to  jetson\_node (\emph{Jetson Nano}). 
Depth matrix data helps to detect the presence of obstacles. Image data feed helps in distinguishing between obstacles and weeds. 

\emph{Arduino Mega} is used to move the rover, equipped with skid steering and comprising four n-20 motors. Through \emph{ROS}-serial communication between \emph{Jetson Nano} and \emph{Arduino Mega}, the former commands the latter. \emph{Jetson Nano} sends the control signal to \emph{Arduino Mega} based on depth values obtained from the camera node (\emph{Zedm}) via the \emph{ROS} framework. If the depth value of the central region is greater than the threshold (here it is set to 0.8m), the vehicle continues to roll on the ground, in the forward direction. The vehicle moves straight, along the direction of the header with a very slight deviation which is adjusted by arranging the landing gear arrangement. In case the value is lesser than the threshold, the vehicle first stops. It then tries to identify if the detected obstacle is a weed or not. This is carried out by analyzing the Point Cloud data received from the image feed of the Camera node (\emph{Zedm}). The data is converted to OpenCV-compatible data using CV Bridge. The RGB data $(r, g, b)$ is then converted to HSV format as follows 

\begin{enumerate}
    \item Divide $r$, $g$, and $b$ by 255.
    \item Compute $c_{\text{max}}$, $c_{\text{min}}$, and $\text{diff}$:
    \begin{itemize}
        \item $c_{\text{max}}$ = $\max(r, g, b)$
        \item $c_{\text{min}}$ = $\min(r, g, b)$
        \item $\text{diff}$ = $c_{\text{max}}$ - $c_{\text{min}}$
    \end{itemize}
    \item Hue is calculated as follows ($h$):
    \[
    \text{if } c_{\text{max}} = c_{\text{min}}, \text{ then } h = 0
    \]
    \[
    \text{if } c_{\text{max}} = r, \text{ then } h = (60 \cdot ((g - b) / \text{diff}) + 360) \% 360
    \]
    \[
    \text{if } c_{\text{max}} = g, \text{ then } h = (60 \cdot ((b - r) / \text{diff}) + 120) \% 360
    \]
    \[
    \text{if } c_{\text{max}} = b, \text{ then } h = (60 \cdot ((r - g) / \text{diff}) + 240) \% 360
    \]
    \item Saturation is calculated as follows ($s$):
    \[
    \text{if } c_{\text{max}} = 0, \text{ then } s = 0
    \]
    \[
    \text{if } c_{\text{max}} \neq 0, \text{ then } s = (\text{diff} / c_{\text{max}}) \times 100
    \]
    \item Value ($v$) is calculated as follows:
    \[
    v = c_{\text{max}} \times 100
    \]
\end{enumerate}

Weed is detected when the HSV data lies in the green colour range. When it does not lie in that range, it indicates the presence of an obstacle and the absence of weed. \textit{Jetson Nano} sends command to \emph{Pixhawk} and enables offboard flight mode. It then sends position setpoints (local coordinate frame) to \emph{Pixhawk} via the mavros node. The vehicle then automatically takes off to a height of 4m and travels a distance of about 2.8m in the forward direction (X direction), and then lands. Upon landing the vehicle continues to roll on the ground in the absence of an obstacle.

The height to which the vehicle has to fly can be determined with the help of the \emph{Zedm} stereo camera. The distance over which the vehicle has to fly can be obtained with the help of a single-point LIDAR. In the case of the prototype developed, LIDAR has not been used and the flight altitude and the distance of flight have been hardcoded to 4m and 2.8m respectively. This is due to computational complexity on account of multiple nodes and serial communication between \emph{Jetson Nano} and \emph{Arduino}, where the latter has a small storage. 

In the case where a major portion of the central region consists of green colour, a command is sent by \emph{Jetson Nano} to \emph{Arduino Mega} through \emph{ROS} framework, to actuate the sprayer arrangement which consists of a nozzle and a small tube connecting it. The spraying process occurs autonomously for 5 seconds on the central region of the crop and then stops. The timing operation is implemented using a 5V relay module. The timing of spraying is adjustable. For the scenario considered, it has been set to 5 seconds. Timing ensures efficient usage of pesticides. Figure \ref{fig:Overall} represents the overall process flow of movement of the hybrid vehicle and spraying activity.
\subsection{Prototype Development}
\subsubsection{Hybrid vehicle with plucker arrangement}
The initial prototype consisted of a mechanical arm integrated to the hybrid drone-rover design. The mechanical arm served as a plucker, aiding in weed removal. The mechanical arm consisted of 3 degrees of freedom and was capable of removing weeds that were 2 to 3 cm thick. Beyond this thickness, sufficient actuation cannot be given for the removal of weed. The base of the mechanical arm is capable of rotating by an angle of about 120 to 140 degrees. Any weed present in this range can be removed by the system. The arm can stretch by a distance of about 10 cm along the horizontal direction. \emph{\emph{\emph{\emph{\emph{Arduino}}}} Mega} was used to controlling the mechanical arm and rover movement via Bluetooth. Wifi can be used to control instead due to better reliability and range. For each DOF, an SG90 servo was used. BO-motors of torque rating of 3.5 kgf. cm and 150 RPM were used for the ground movement of the vehicle.
\subsubsection{Hybrid vehicle with sprayer arrangement}
\begin{figure}[htp]
\begin{center}
\includegraphics[scale=0.50]{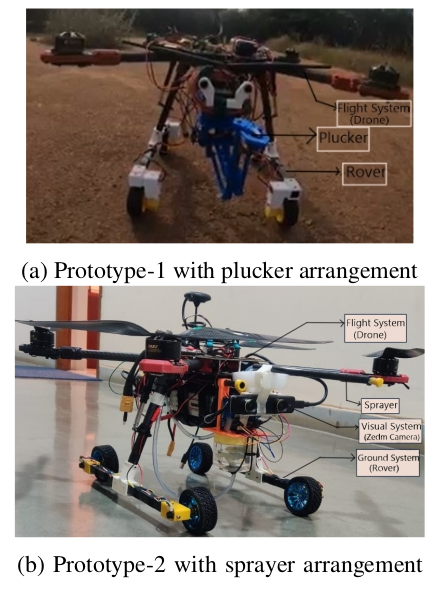}
\end{center}
\caption{Figure (a) represents the prototype of the hybrid vehicle with a plucker arrangement. Figure (b) represents the prototype of the hybrid vehicle with sprayer arrangement.}
\label{fig_plucker_arrangement}
\end{figure}

The second prototype consists of a sprayer system integrated with the hybrid vehicle, which consists of \emph{Jetson Nano}, \emph{\emph{\emph{\emph{\emph{Arduino}}}} Mega}, \emph{Zed Mini} (\emph{Zedm}) Camera, and \emph{Pixhawk}. Serial communication between Nvidia \emph{Jetson Nano} and \emph{\emph{\emph{\emph{\emph{Arduino}}}} Mega} through the \emph{ROS} framework guides the ground movement of the vehicle along with the action of spraying. \emph{Pixhawk} and \emph{Jetson Nano} aid in the flight of the system. The timing of the spraying actuation is controlled with the help of a 5V relay module. Spraying actuation on the weeds is carried out with the help of a 5V submersible pump with a maximum flow rate of 120 Liters Per Hour. Spraying on weeds can be carried out from a distance of about 1m. Figure \ref{fig_plucker_arrangement} shows the prototypes developed. Prototype 1 refers to the prototype with a plucker arrangement and prototype 2 is of sprayer arrangement.

%

\section{Experimental Results}
In the experimental setup considered, the vehicle is placed at a distance of about 2 to 2.5m from an obstacle, represented by the blue tarpaulin sheet. From the tarpaulin sheet, there is a pot placed at a distance of about 2.5m. The pot and the tarpaulin sheet are placed along the direction of the header of the hybrid drone-rover system. The blue tarpaulin sheet symbolizes a wall of height 3m and a width of 1.2m. The pot consists of a plant that symbolizes weeds in agricultural farms. The tarpaulin sheet can be visualized as obstacles or canals in farmlands and as racks in vertical farms.

The section briefs about the results achieved in the testing of the prototype. It includes hybrid vehicle movement, obstacle detection, crop detection, and tackling weeds via spraying and plucker arrangement.
\subsubsection{Obstacle Detection and Flight mode}
\begin{figure}[htp]
\begin{center}
\includegraphics[scale=0.50]{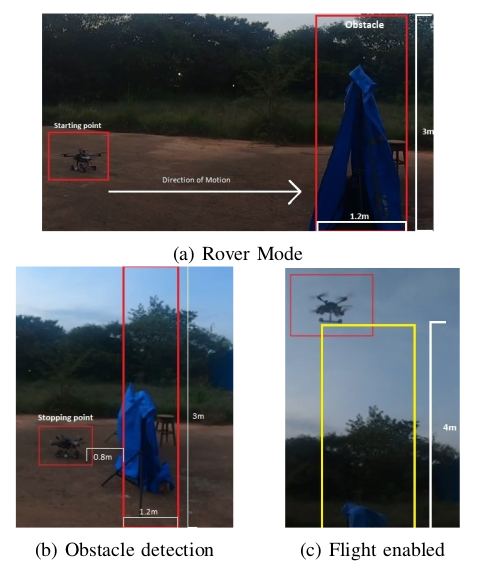}
\end{center}
\caption{Figure (a) represents operation in rover mode. Figure (b) illustrates obstacle detection (rover stops upon detecting obstacle). Figure (c) illustrates flight mode enabled upon obstacle detection.}
\label{fig_rover}
\end{figure}

The hybrid vehicle keeps rolling on the ground in the forward direction. It stops when it detects an object, in our case the blue tarpaulin sheet. It stops when it is around a distance of about 0.8m from it. The vehicle switches to flight mode. It then takes off to an altitude of about 4m and crosses the object of height 3m. It moves about 2.8m in the forward direction with an error deviation of less than 0.1m, which is associated with external parameters such as wind. It crosses the obstacle of width 1.2m and then lands. Upon landing, the vehicle continues to move on the ground in the forward direction, till the pot is detected.
Figure \ref{fig_rover}, part a describes the ground movement of the hybrid drone-rover vehicle. Part b represents the hybrid vehicle halting upon detecting the obstacle (tarpaulin sheet), before it switches to flight mode as described in part c, to move over it.

\subsubsection{Detection and Spraying Actuation}
\begin{figure}[htp]
\begin{center}
\includegraphics[scale=0.250]{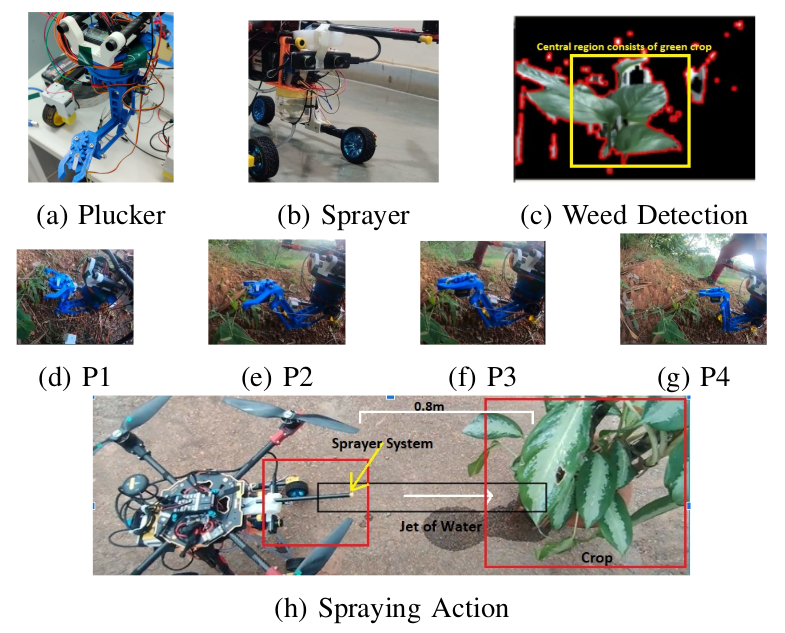}
\end{center}
\caption{Figures (a) and (b) represent the plucker and sprayer arrangement. Figure (c) describes the output of weed detection. Figures (d) to (g) illustrate the process of weed removal using the plucker arrangement. Figure (h) illustrates the process of spraying activity.}
\label{fig_plu_rov}
\end{figure}

Upon landing, the system continues to roll on the ground till the pot is detected. The rover rolls till it reaches close to the pot(about 0.8m). The plant is then detected and spraying is carried out on the central portion of the plant from a distance of about 0.8m for a couple of seconds and then stopped. Figure \ref{fig_plu_rov}, parts a and b represent the plucker and sprayer arrangement. Part c describes the output of weed detection. Parts d to g illustrate the process of weed removal using the plucker arrangement. Part h illustrates the process of spraying activity.

\subsubsection{Prototype Validation}
In the prototype developed, a compact hybrid vehicle has been designed and its nature verified. Obstacle detection has been implemented. Modularity in weed removal system designs has been seen. Thus, concerning the needs of the customer, functional requirements have been satisfied.
\subsubsection{Cost Analysis}
Currently, drones are being used more extensively in agriculture than rovers due to their long-range capability. The cost mentioned below is as per available internet sources. Typical agricultural drones used for spraying fertilizers and pesticides are of frame size 960mm and greater and range from 4,00,000 \rupee(INR) onwards. Rovers for seeding application cost around 2,50,000 \rupee(INR) and for weed removal application is seen to be more than that. The developed prototype is estimated to cost  1,00,000 \rupee(INR) as it uses a smaller drone frame of size 650mm. The estimated cost is around one-fourth of the cost of a regular agricultural drone of frame size 960mm. 
\begin{table}[!h]
\caption{Cost Analysis of Developed Prototype}
\label{tab:template}
\centering
\begin{tabular}{ |p{5cm}|p{2cm}|}
\hline
\centering Feature & Cost (\rupee:INR) \\
\hline
 Hybrid Vehicle Mechanical Design & 15,000  \\
 \hline
 Hardware Of System & 80,000  \\
 \hline
 Plucker and Sprayer Mechanical Design & 5,000 \\
\hline
\end{tabular}

\end{table}
 \section{Conclusion}
This paper proposes and successfully implements a novel prototype of hybrid vehicle with weed removal capability, with the help of the plucker and sprayer arrangement respectively. Rover mode helps to tackle the weeds effectively at the stem level. Detection is successfully carried out with the help of the visual system using the HSV algorithm.  Ground and flight movement has been successfully seen. Obstacle detection along with the automatic switch between drone and rover mode is also demonstrated. This helps the vehicle to move across canals and uneven terrains in agricultural fields. A compact design is developed which helps in easy movement between rows of crops. The key purpose of this paper is to develop a proof of concept, capable of meeting the basic functional requirements such as ground and flight mobility,  automatic switch between drone-rover capability, and weed detection and removal capability, The basic functional requirements have been seen as described above. 


\bibliography{mybibfile}

\end{document}